\documentclass[letterpaper, 10 pt, conference]{ieeeconf}  % Comment this line out if you need a4paper

\IEEEoverridecommandlockouts                              % This command is only needed if 
\usepackage{multicol}
\usepackage[bookmarks=true]{hyperref}

\usepackage{algorithm2e}
\usepackage[english]{babel}
\usepackage{color}
\usepackage{epic}
\usepackage{epsfig}
\usepackage{epstopdf}
\usepackage[multiple]{footmisc}
\usepackage{graphicx}
\graphicspath{{figures/}}
\usepackage{placeins}
\usepackage{wrapfig}
\usepackage[dvipsnames]{xcolor}

\usepackage{array}
\newcolumntype{M}[1]{>{\centering\arraybackslash}m{#1}}
\newcolumntype{N}{@{}m{0pt}@{}}

\title{\LARGE \bf
PRIMAL: Pathfinding via Reinforcement and Imitation \\ Multi-Agent Learning
}

\author{Guillaume Sartoretti$^{1}$, Justin Kerr$^{1}$, Yunfei Shi$^{1}$, Glenn Wagner$^{2}$, \\ T. K. Satish Kumar$^{3}$, Sven Koenig$^{3}$, and Howie Choset$^{1}$% <-this % stops a space
\thanks{G. Sartoretti, H. Choset, J. Kerr, Y. Shi are with the Robotics Institute at Carnegie Mellon University, Pittsburgh, PA 15213, USA. {\tt\small \{gsartore,jgkerr,yunfeischoset\}@andrew.cmu.edu}.\protect\\
G. Wagner is with the Commonwealth Scientific and Industrial Research Organisation (CSIRO), Pullenvale QLD 4069, Australia, {\tt\small glenn.s.wagner@gmail.com}.\protect\\
T. K. S. Kumar and S. Koenig are with the Computer Science Department at the University of Southern California, Los Angeles, CA 90089, USA. {\tt\small tkskwork@gmail.com, skoenig@usc.edu}.}% <-this % stops a space
}

\begin{document}

\maketitle
\thispagestyle{empty}
\pagestyle{empty}

%%%%%%%%%%%%%%%%%%%%%%%%%%%%%%%%%%%%%%%%%%%%%%%%%%%%%%%%%%%%%%%%%%%%%%%%%%%%%%%%
%%%%%%%%%%%%%%%%%%%%%%%%%%%%%%%%%%%%%%%%%%%%%%%%%%%%%%%%%%%%%%%%%%%%%%%%%%%%%%%%

\begin{abstract}

% 199 words, 200 limit!!
Multi-agent path finding (MAPF) is an essential component of many large-scale, real-world robot deployments, from aerial swarms to warehouse automation. However, despite the community's continued efforts, most state-of-the-art MAPF planners still rely on centralized planning and scale poorly past a few hundred agents. Such planning approaches are maladapted to real-world deployments, where noise and uncertainty often require paths be recomputed online, which is impossible when planning times are in seconds to minutes. We present PRIMAL, a novel framework for MAPF that combines reinforcement and imitation learning to teach fully-decentralized policies, where agents reactively plan paths online in a partially-observable world while exhibiting implicit coordination. This framework extends our previous work on distributed learning of collaborative policies by introducing demonstrations of an expert MAPF planner during training, as well as careful reward shaping and environment sampling. Once learned, the resulting policy can be copied onto any number of agents and naturally scales to different team sizes and world dimensions. We present results on randomized worlds with up to 1024 agents and compare success rates against state-of-the-art MAPF planners. Finally, we experimentally validate the learned policies in a hybrid simulation of a factory mockup, involving both real-world and simulated robots.

\end{abstract}

%%%%%%%%%%%%%%%%%%%%%%%%%%%%%%%%%%%%%%%%%%%%%%%%%%%%%%%%%%%%%%%%%%%%%%%%%%%%%%%%
%%%%%%%%%%%%%%%%%%%%%%%%%%%%%%%%%%%%%%%%%%%%%%%%%%%%%%%%%%%%%%%%%%%%%%%%%%%%%%%%

\section{INTRODUCTION}
\label{RAL2018-introduction}

\begin{figure}[t]
\vspace{0.15cm}
\begin{center}
\includegraphics[width=\linewidth]{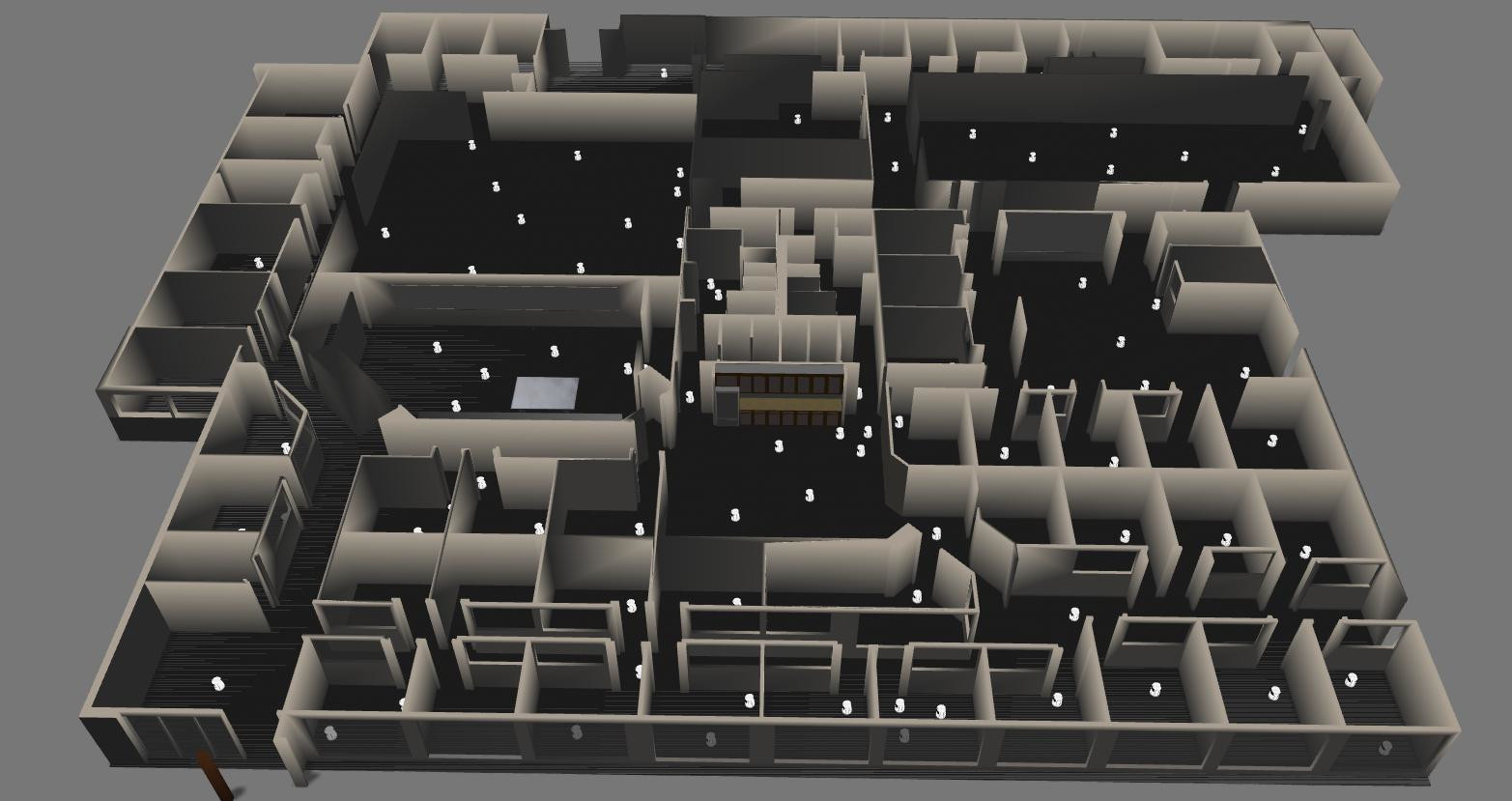}
\end{center}
\vspace{-0.5cm}
\caption{Example problem where $100$ simulated robots (white dots) must compute individual collision-free paths in a large, factory-like environment.}
\label{RAL2018-fig:frontFigure}
\vspace{-0.4cm}
\end{figure}

Given the rapid development of affordable robots with embedded sensing and computation capabilities,
% we are quickly approaching a point at which
manufacturing applications will soon regularly involve the deployment of thousands of robots~\cite{Rubenstein2014,Howard2006}.
To support these applications, significant research effort has been devoted to multi-agent path finding (MAPF)~\cite{wagner2015subdimensional,Barer2014,van2011reciprocal,Cui2011} for deployment in distribution centers and potential use for airplane taxiing~\cite{baxter2007multi,balakrishnan2007framework}.
However, as the number of agents in the system grows, so does the complexity of coordinating them.
Current state-of-the-art optimal planners can plan for several hundreds of agents, and the community is now settling for bounded suboptimal planners as a potential solution for even larger multi-agent systems~\cite{wagner2015subdimensional,wang2011mapp}.
Another common approach is to rely on reactive planners, which do not plan joint paths for all agents before execution, but rather correct individual paths online to avoid collisions~\cite{van2011reciprocal,chen2017decentralized}.
However, such planners often prove inefficient in cluttered factory environments (such as Fig.~\ref{RAL2018-fig:frontFigure}), where they can result in dead- and livelocks~\cite{van2011reciprocal}.

% As a new direction for research in multi-agent path planning,
% for large numbers of robots, we propose to build upon recent results in distributed reinforcement learning (RL)~\cite{sutton1998reinforcement} to train
% We propose to approach the MAPF problem from the standpoint of distributed reinforcement learning (RL) to teach each agent to plan its path online.
Extending our previous work on distributed reinforcement learning (RL) for multiple agents in shared environments~\cite{DARS2018-DistributedAssembly,ICRA2018-DistributedLearning}, the main contribution of this paper introduces PRIMAL, a novel hybrid framework for decentralized MAPF that combines RL~\cite{sutton1998reinforcement} and imitation learning (IL) from an expert centralized MAPF planner.
In this framework, agents learn to take into account the consequences of their position on other agents, in order to favor movements that will benefit the whole team and not only themselves.
That is, by simultaneously learning to plan efficient single-agent paths (mostly via RL), and to imitate a centralized expert (IL), agents ultimately learn a decentralized policy where they still exhibit implicit coordination during online path planning without the need for explicit communication among agents.
Since multiple agents learn a common, single-agent policy, the final learned policy can be copied onto any number of agents.
Additionally, we consider the case where agents evolve in a partially-observable world, where they can only observe the world in a limited field of view (FOV) around themselves.
We present the results of an extensive set of simulation experiments and show that the final, trained policies naturally scale to various team and world sizes.
We further highlight cases where PRIMAL outperforms other state-of-the-art MAPF planners and cases where it struggles. % (optimal/suboptimal, centralized/decentralized).
We also present experimental results of the trained policy in a hybrid simulation of a factory mockup.

The paper is structured as follows:
In Section~\ref{RAL2018-background}, we summarize the state-of-the-art in MAPF and multi-agent RL.
We detail how MAPF is cast in the RL framework in Section~\ref{RAL2018-policy}, and how learning is carried out in Section~\ref{RAL2018-learning}.
% Section~\ref{RAL2018-results} presents our numerical and experimental results, while we provide concluding remarks in Section~\ref{RAL2018-conclusion}.
Section~\ref{RAL2018-results} presents our results, and Section~\ref{RAL2018-conclusion} concluding~remarks.

%%%%%%%%%%%%%%%%%%%%%%%%%%%%%%%%%%%%%%%%%%%%%%%%%%%%%%%%%%%%%%%%%%%%%%%%%%%%%%%%
%%%%%%%%%%%%%%%%%%%%%%%%%%%%%%%%%%%%%%%%%%%%%%%%%%%%%%%%%%%%%%%%%%%%%%%%%%%%%%%%

\section{PRIOR WORK}
\label{RAL2018-background}

%%%%%%%%%%%%%%%%%%%%%%%%%%%%%%%%%%%%%%%%%%%%%%%%%%%%%%%%%%%%%%%%%%%%%%%%%%%%%%%

\subsection{Multi-Agent Path Finding (MAPF)}
\label{RAL2018-background-MAPF}

MAPF %is a central problem in robotics, and
is an NP-hard problem even when approximating optimal solutions~\cite{ma2019prioritization,LaValle_2006}.
MAPF planners can be broadly classified into three categories: coupled, decoupled, and dynamically-coupled approaches.
Coupled approaches (e.g., standard $A^*$), which treat the multi-agent system as a single, very high dimensional agent, greatly suffer from an exponential growth in planning complexity.
Hence we focus on decoupled and dynamically-coupled, state-of-the-art planners for large MAPF problems.

Decoupled approaches compute individual paths for each agent, and then adjust these paths to avoid collisions.
Since individual paths can be planned, as well as adjusted for collisions, in low-dimensional search spaces, decoupled approaches can rapidly find paths for large multi-agent systems~\cite{van2011reciprocal,Leroy_1999}.
Velocity planners fix the individual path that will be followed by each agent, then find a velocity profile along those paths that avoids collisions~\cite{Cui2011,chen2017decentralized}.
In particular, ORCA~\cite{van2011reciprocal} adapts the agents' velocity magnitudes and directions online to avoid collisions, on top of individually-planned single-agent paths, and recent work has focused on such an obstacle avoidance approach using reinforcement learning (RL)~\cite{chen2017decentralized}.
Priority planners assign a priority to each agent, and plan individual paths in decreasing order of priority, each time treating higher priority agents as moving obstacles~\cite{ma2016multi,Cap2013prioirty,Erdmann_1987}.
The main drawback of decoupled approaches is that the low-dimensional search spaces used only represent a small portion of the joint configuration space, meaning that these approaches cannot be \textit{complete} (i.e., find paths for all solvable problems)~\cite{Sanchez_2002_centralized}.

% 2.3. Hybrid multi-agent path planning
Several recent approaches lie between coupled and decoupled approaches: they allow for richer agent-agent behaviors than can be achieved with decoupled planners, while avoiding planning in the joint configuration space.
A common approach followed by dynamically coupled approaches is to grow the search space as necessary during planning~\cite{wagner2015subdimensional,sharon2012conflict}.
Conflict-Based Search (CBS) and its variants~\cite{Barer2014,sharon2012conflict} plans for individual agents and constructs a set of constraints to find optimal or near-optimal solutions without exploring higher-dimensional spaces.
Extending standard $A^*$ to MAPF, $M^*$ and its variants~\cite{wagner2015subdimensional} first plan paths for individual agents and then project these individual plans forward through time searching for collisions.
The configuration space is only locally expanded around any collision between single-agent plans, where joint planning is performed through (usually limited) backtracking to solve the collision and resume single-agent plans.
In particular, OD-recursive-$M^*$ (ODrM*)~\cite{Ferner_2013} can further reduce the set of agents for which joint planning is necessary, by breaking it down into independent collision sets, combined with Operator Decomposition (OD)~\cite{Standley_2010} to keep the branching factor small during search.

%%%%%%%%%%%%%%%%%%%%%%%%%%%%%%%%%%%%%%%%%%%%%%%%%%%%%%%%%%%%%%%%%%%%%%%%%%%%%%%

\subsection{Multi-Agent Reinforcement Learning (MARL)}
\label{RAL2018-background-MARL}

The first and most important problem encountered when transitioning from single- to multi-agent learning is the curse
of dimensionality: most joint approaches fail as the state-
action spaces explode combinatorially, requiring impractical amounts of training data to converge~\cite{Buoniu2010}.
In this context, many recent work have focused on decentralized policy learning~\cite{gupta2017cooperative,lowe2017multi,foerster2017counterfactual,foerster2016learning,melo2013heuristic}, where agents each learn their own policy, which should encompass a measure of agent cooperation, at least during training.
One such approach is to train agents to predict other agents' actions~\cite{lowe2017multi,foerster2017counterfactual}, which generally scales poorly as the team size increases.
In most cases, some form of centralized learning is involved, where the sum of experience of all agents can be used towards training a common aspect of the problem (e.g., network output or value/advantage calculation)~\cite{gupta2017cooperative,foerster2017counterfactual,foerster2016learning}.
When centrally learning a network output, parameter sharing has been used to enable faster and more stable training by sharing the weights of some of the layers of the neural net~\cite{gupta2017cooperative}.
In actor-critic approaches, for example, the critic output of the network is often trained centrally with parameter sharing, since it applies to all agents in the system, and has been used to train cooperation between agents~\cite{gupta2017cooperative,foerster2017counterfactual}.
Centralized learning can also help when dealing with partially-observable systems, by aggregating all the agents' observations into a single learning process~\cite{gupta2017cooperative,foerster2017counterfactual,foerster2016learning}.

Second, many existing approaches rely on explicit communication among agents, to share observations or selected actions during training and sometimes also during policy execution~\cite{lowe2017multi,foerster2017counterfactual,foerster2016learning}.
In our previous work~\cite{DARS2018-DistributedAssembly,ICRA2018-DistributedLearning}, we focused on extending the state-of-the-art asynchronous advantage actor-critic (A3C) algorithm to enable multiple agents to learn a common, homogeneous policy in shared environments without the need for any explicit agent communication.
That is, the agents had access to the full state of the system (fully-observable world), and treated each other as moving obstacles.
There, stabilizing learning is key: the learning gradients obtained by agents experiencing the same episode in the same environment are often very correlated and destabilized the learning process.
To prevent this, we relied on experience replay~\cite{experience} and carefully randomized episode initialization.
However, we did not train agents to exhibit any form of coordination.
That is, in our previous extension of A3C, agents \textbf{collaborate} (i.e., work towards a common goal) but do not explicitly \textbf{cooperate} (i.e., take actions to benefit the whole group and not only themselves).

In our work, we propose to rely on imitation learning (IL)of an expert centralized planner (ODrM*) to train agents to exhibit coordination, without the need for explicit communication, in a partially-observable world.
We also propose a carefully crafted reward structure and a way to sample the challenges used to train the agents.
The resulting, trained policy is executed by each agent based on locally gathered information but still allows agents to exhibit cooperative behavior, and is also robust against agent failures or additions.

%%%%%%%%%%%%%%%%%%%%%%%%%%%%%%%%%%%%%%%%%%%%%%%%%%%%%%%%%%%%%%%%%%%%%%%%%%%%%%%%
%%%%%%%%%%%%%%%%%%%%%%%%%%%%%%%%%%%%%%%%%%%%%%%%%%%%%%%%%%%%%%%%%%%%%%%%%%%%%%%%

\section{POLICY REPRESENTATION}
\label{RAL2018-policy}

In this section, we present how the MAPF problem is cast into the RL framework.
We detail the observation and action spaces of each agent, the reward structure and the neural network that represents the policy to be learned.

%%%%%%%%%%%%%%%%%%%%%%%%%%%%%%%%%%%%%%%%%%%%%%%%%%%%%%%%%%%%%%%%%%%%%%%%%%%%%%%%

\subsection{Observation Space}
\label{RAL2018-policy-stateSpace}

\begin{figure}[t]
\vspace{0.2cm}
\begin{center}
\includegraphics[width=0.88\linewidth]{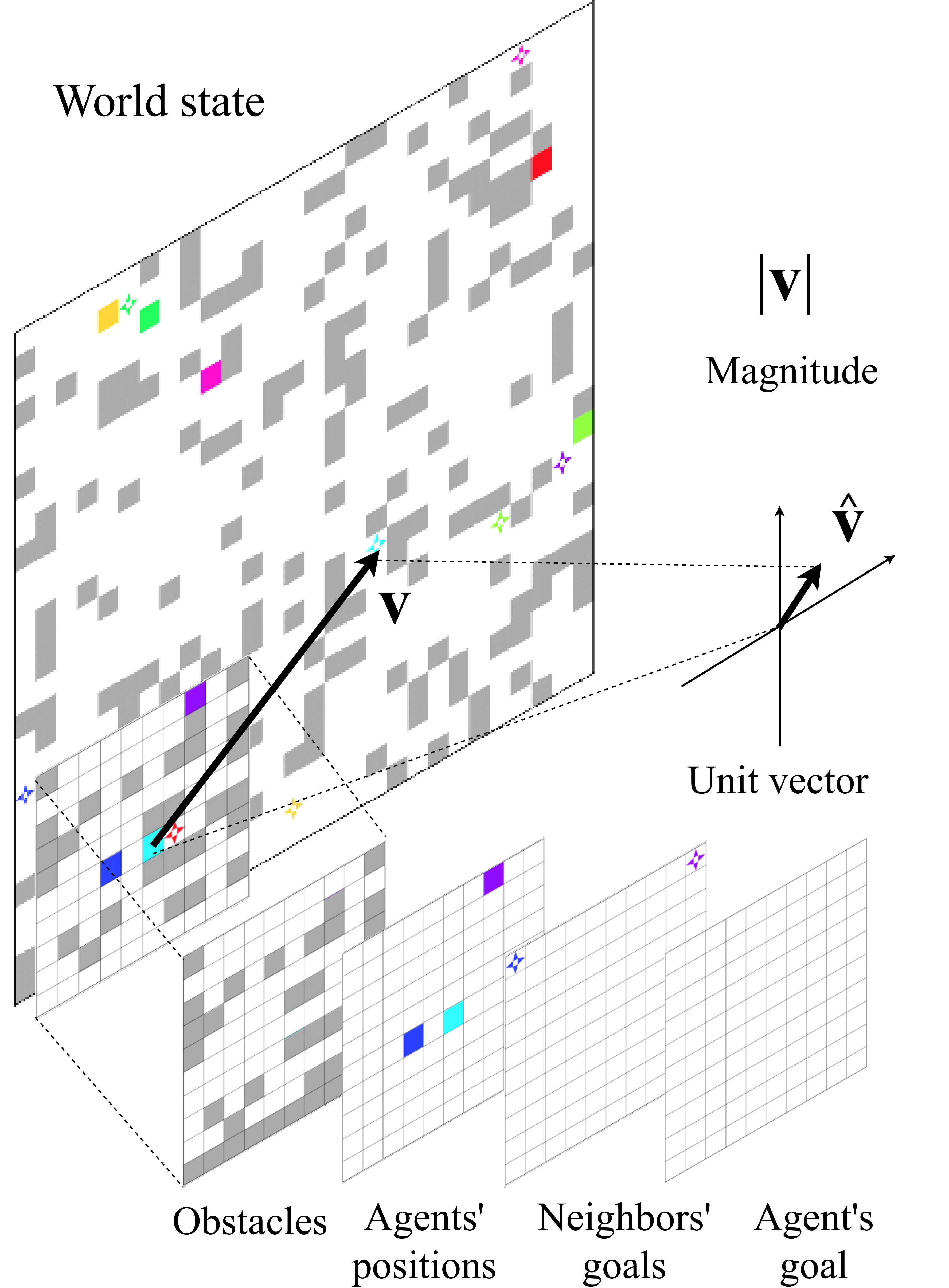}
\end{center}
\vspace{-0.6cm}
\caption{Observation space of each agent (here, for the light blue agent).
Agents are displayed as colored squares, their goals as similarly-colored stars, and obstacles as grey squares.
Each agent only has access to a limited field of view (FOV) centered around its position, in which information is broken down into channels: positions of obstacles, position of nearby agents, goal positions of these nearby agents (projected onto the boundary of the FOV if outside of the FOV), and position of its goal if within the FOV.
Note how the bottom row of the obstacle channel has been filled with obstacles, since these positions are outside of the world's boundaries.
Each agent also has access to a normalized vector pointing to its goal (often outside of its FOV) and its magnitude (distance to goal), as a natural way to let agents learn to select their general direction of travel.}
\label{RAL2018-fig:observation}
\vspace{-0.3cm}
\end{figure}

We consider a partially-observable discrete gridworld, where agents can only observe the state of the world in a limited FOV centered around themselves ($10\times10$ FOV in practice).
We believe that considering a partially-observable world is an important step towards real-world robot deployment.
In scenarios where the full map of the environment is available (e.g., automated warehouses), it is always possible to train agents with full observability of the system by using a sufficiently large FOV.
Additionally, assuming a fixed FOV can allow the policy to generalize to arbitrary world sizes and also helps to reduce the input dimension to the neural network.
However, an agent needs to have access to information about its goal, which is often outside of its FOV.
To this end, it has access to both a unit vector pointing towards its goal and Euclidean distance to its goal at all times (see Figure~\ref{RAL2018-fig:observation}).

In the limited FOV, we separate the available information into different channels to simplify the agents' learning task.
Specifically, each observation consists of binary matrices representing the obstacles, the positions of other agents, the agent's own goal location (if within the FOV), and the position of other observable agents' goals.
When agents are close to the edges of the world, obstacles are added at all positions outside the world's boundaries.% in the obstacle channel.

%%%%%%%%%%%%%%%%%%%%%%%%%%%%%%%%%%%%%%%%%%%%%%%%%%%%%%%%%%%%%%%%%%%%%%%%%%%%%%%%

\subsection{Action Space}
\label{RAL2018-policy-actionSpace}

Agents take discrete actions in the gridworld: moving one cell in one of the four cardinal directions or staying still.
At each timestep, certain actions may be invalid, such as moving into a wall or another agent.
During training, actions are sampled only from valid actions and an additional loss function aids in learning this information.
We experimentally observed that this approach enables more stable training, compared to giving negative rewards to agents for selecting invalid moves.
Additionally, to combat convergence to oscillating policies, agents are prevented during training from returning to the location they occupied at the last timestep (agents can still stay still during multiple successive timesteps).
This is necessary to encourage exploration and learn effective policies (even when also using IL).

If an agent selects an invalid move during testing, it instead stays still for that timestep.
In practice, agents very rarely select invalid moves once fully trained, showing that they effectively learn the set of valid actions in each state.

\begin{figure*}[t]
\vspace{0.1cm}
\begin{center}
\includegraphics[width=\linewidth]{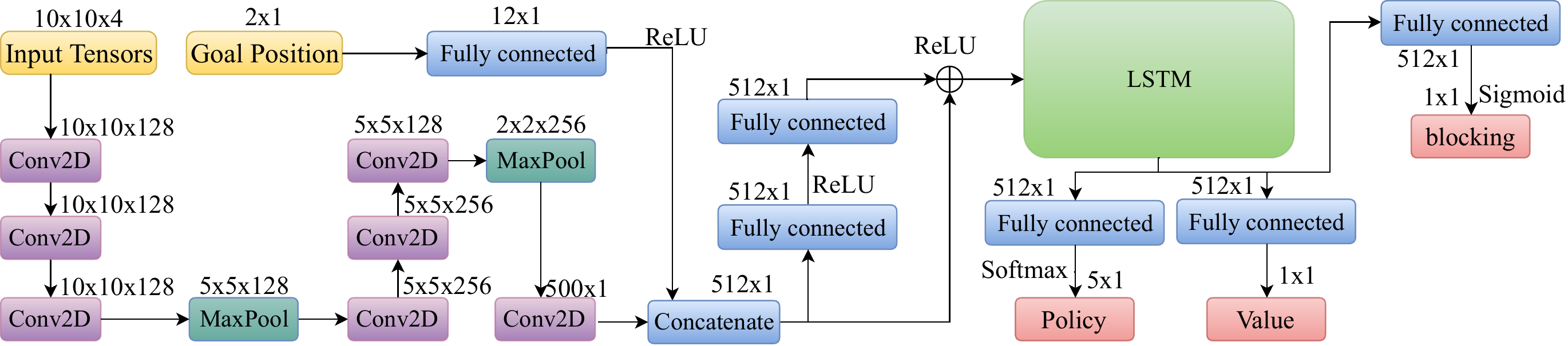}
\end{center}
\vspace{-0.5cm}
\caption{The neural network consists of $7$ convolutional layers interleaved with maxpooling layers, followed by an LSTM.}
\label{RAL2018-fig:networkDiagram}
\vspace{-0.3cm}
\end{figure*}

%%%%%%%%%%%%%%%%%%%%%%%%%%%%%%%%%%%%%%%%%%%%%%%%%%%%%%%%%%%%%%%%%%%%%%%%%%%%%%%%

\subsection{Reward Structure}
\label{RAL2018-policy-rewards}

Our reward function (Table~\ref{RAL2018-tab:rewardFunction}) follows the same intuition that most reward functions for gridworlds use, where agents are punished for each timestep they are not resting on goal, leading to the strategy of reaching their goals as quickly as possible. %~\cite{we should be able to find some sources using this}.
We penalize agents slightly more for staying still than for moving, which is necessary to encourage exploration.
Even though imitation assists in exploration, we found that removing this aspect of the reward function led to poor convergence, which might be the case due to conflicts between the RL and IL gradients.
Though invalid moves (moving back to the previous cell, or into an obstacle) are filtered out of the action space during training as described in Section~\ref{RAL2018-policy-actionSpace} because agents act sequentially in a random order, it is still possible for them to collide, e.g., when multiple agents choose to move to the same location at the same timestep.
Agent collisions result in a $-2$ reward.
Agents receive a $+20$ reward for finishing an episode, i.e., when all agents are on their goals simultaneously.
% Our reward structure is summarized in~Table~\ref{RAL2018-tab:rewardFunction}.

\vspace{-.2cm}
\begin{table}[h]
\caption{Simple reward structure.}
\label{RAL2018-tab:rewardFunction}
\vspace{-.5cm}
\begin{center}
\begin{tabular}{|M{5cm}|M{1.66cm}|N}
\hline
\textbf{Action} & \textbf{Reward} &\\[0.15cm]
\hline
Move [N/E/S/W] & -0.3 &\\[0.15cm]
\hline
Agent Collision & -2.0 &\\[0.15cm]
\hline
No Movement (on/off goal) & 0.0 / -0.5 &\\[0.15cm]
\hline
Finish Episode & +20.0 &\\[0.15cm]
\hline
\end{tabular}
\end{center}
\vspace{-.5cm}
\end{table}

%%%%%%%%%%%%%%%%%%%%%%%%%%%%%%%%%%%%%%%%%%%%%%%%%%%%%%%%%%%%%%%%%%%%%%%%%%%%%%%%

\vspace{-.22cm}
\subsection{Actor-Critic Network}
\label{RAL2018-policy-ACNet}

Our work relies on the asynchronous advantage actor-critic (A3C) algorithm~\cite{mnih2016asynchronous} and extends our previous work
on distributed learning for multiple agents in shared environments~\cite{DARS2018-DistributedAssembly,ICRA2018-DistributedLearning}.
We use a deep neural network to approximate the agent's policy, which maps the current observation of its surroundings to the next action to take.
This network has multiple outputs, one of them being the actual policy and the others only being used toward training it.
We use the 6-layer convolutional network pictured in Fig.~\ref{RAL2018-fig:networkDiagram}, taking inspiration from VGGnet~\cite{simonyan2014very}, using several small $3\times3$ kernels between each max-pooling layer.

Specifically, the two inputs to the neural network -- the local observation and the goal direction/distance -- are pre-processed independently, before being concatenated half-way through the neural network.
The four-channel matrices ($10 \times 10 \times 4$ tensor) representing the local observation are passed through two stages of three convolutions and maxpooling, followed by a last convolutional layer.
In parallel, the goal unit vector and magnitude are passed through one fully-connected (fc) layer.
The concatenation of both of these pre-processed inputs is then passed through two fc layers, which is finally fed into a long-short-term memory (LSTM) cell with output size $512$.
A residual shortcut~\cite{resnet} connects the output of the concatenation layer to the input layer of the LSTM.
The output layers consist of the policy neurons with softmax activation, the value output,
and a feature layer used to train each agent to know whether it is blocking other agents from reaching their goals (detailed in Section~\ref{RAL2018-learning-blocking}).

During training, the policy, value, and ``blocking'' outputs are updated in batch every $n = 256$ steps or when an episode finishes.
As is common, the value is updated to match the total discounted return ($R_t=\sum_{i=0}^k\gamma^i r_{t+i}$) by minimizing:

\vspace{-.3cm}
\begin{equation}
L_{V}=\sum_{t=0}^T (V(o_t;\theta)-R_t)^2.
\label{RAL2018-valueloss}
\end{equation}
\vspace{-.38cm}

To update the policy, we use an approximation of the advantage function by bootstrapping using the value function:
$A(o_t, a_t; \theta) = \sum_{i=0}^{k-1} \gamma^i r_{t+i} + \gamma^k V(o_{k+t}; \theta) - V(o_t; \theta)$ (where $k$ is bounded by the batch size $T$).
We also add an entropy term $H(\pi(o))$ to the policy loss, which has been shown to encourage exploration and discourage premature convergence~\cite{babaeizadeh2016reinforcement} by penalizing a policy that always chooses
the same actions. The policy loss reads

\vspace{-.55cm}
\begin{equation}
\label{RAL2018-policyloss}
L_{\pi} = \sigma_H \cdot H(\pi(o)) - \sum_{t=0}^T \log(P(a_t | \pi, o; \theta) A(o_t, a_t; \theta))
\end{equation}
\vspace{-.35cm}

\noindent with a small entropy weight $\sigma_H$ ($\sigma_H = 0.01$ in practice).
We rely on two additional loss functions which help to guide and stabilize training.
First, the blocking prediction output is updated by minimizing $L_{blocking}$, the log likelihood of
% predicting incorrectly. Second, we define the loss function
predicting incorrectly. Second, we define the loss function $L_{valid}$ to minimize the log likelihood of selecting an invalid move~\cite{DARS2018-DistributedAssembly}, as mentioned in Section~\ref{RAL2018-policy-actionSpace}.

%%%%%%%%%%%%%%%%%%%%%%%%%%%%%%%%%%%%%%%%%%%%%%%%%%%%%%%%%%%%%%%%%%%%%%%%%%%%%%%%
%%%%%%%%%%%%%%%%%%%%%%%%%%%%%%%%%%%%%%%%%%%%%%%%%%%%%%%%%%%%%%%%%%%%%%%%%%%%%%%%

\section{LEARNING}
\label{RAL2018-learning}

In this section, we detail our distributed framework for learning MAPF with implicit agent coordination.
The RL portion of our framework builds upon our previous work on distributed RL for multiple agents in shared environments~\cite{DARS2018-DistributedAssembly,ICRA2018-DistributedLearning}.
In our work, we introduce an IL module that allows agents to learn from expert demonstrations.

%%%%%%%%%%%%%%%%%%%%%%%%%%%%%%%%%%%%%%%%%%%%%%%%%%%%%%%%%%%%%%%%%%%%%%%%%%%%%%%%

\subsection{Coordination Learning}
\label{RAL2018-learning-CL}

One of the key challenges in training a decentralized policy is to encourage agents to act selflessly, even though it may be detrimental to their immediate maximization of reward.
In particular, agents typically display undesirable selfish behavior when stopped on their goals while blocking other agents' access to their own goals.
A naive implementation of our previous work~\cite{DARS2018-DistributedAssembly}, where agents distributedly learn a fully selfish policy, fails in dense environments with many narrow environmental features where the probability of blocking other agents is high.
That is, agents simply learn to move as fast as possible to their goals, and then to never move away from it, not even to let other agents access their own goals (despite the fact that this would end the episode earlier, which would result in higher rewards for all agents).

Many of the current multi-agent training techniques addressing this selfishness problem are invalidated by the size of the environments and the limited FOV of agents.
Shared critics~\cite{foerster2017counterfactual} have proven effective at multi-agent credit assignment.
However, these methods are typically used when agents have almost full information about their environment.
In our highly decentralized scenario, assigning credit to agents may be confusing when they cannot observe the source of the penalty, for example, when an agent cannot observe that a long hallway is a dead-end, yet the universal critic sharply decreases the value function.
Another popular multi-agent training technique is to apply joint rewards to agents in an attempt to help them realize the benefit of taking personal sacrifices to benefit the team~\cite{ying2008improvement,ICRA2018-DistributedLearning}.
We briefly tried to assign joint rewards to agents within the same FOV.
However, this produced no noticeable difference in behavior, so we abandoned it in favor of the methods described below.

\begin{figure}[t]
\vspace{0.2cm}
\begin{center}
\includegraphics[width=\linewidth]{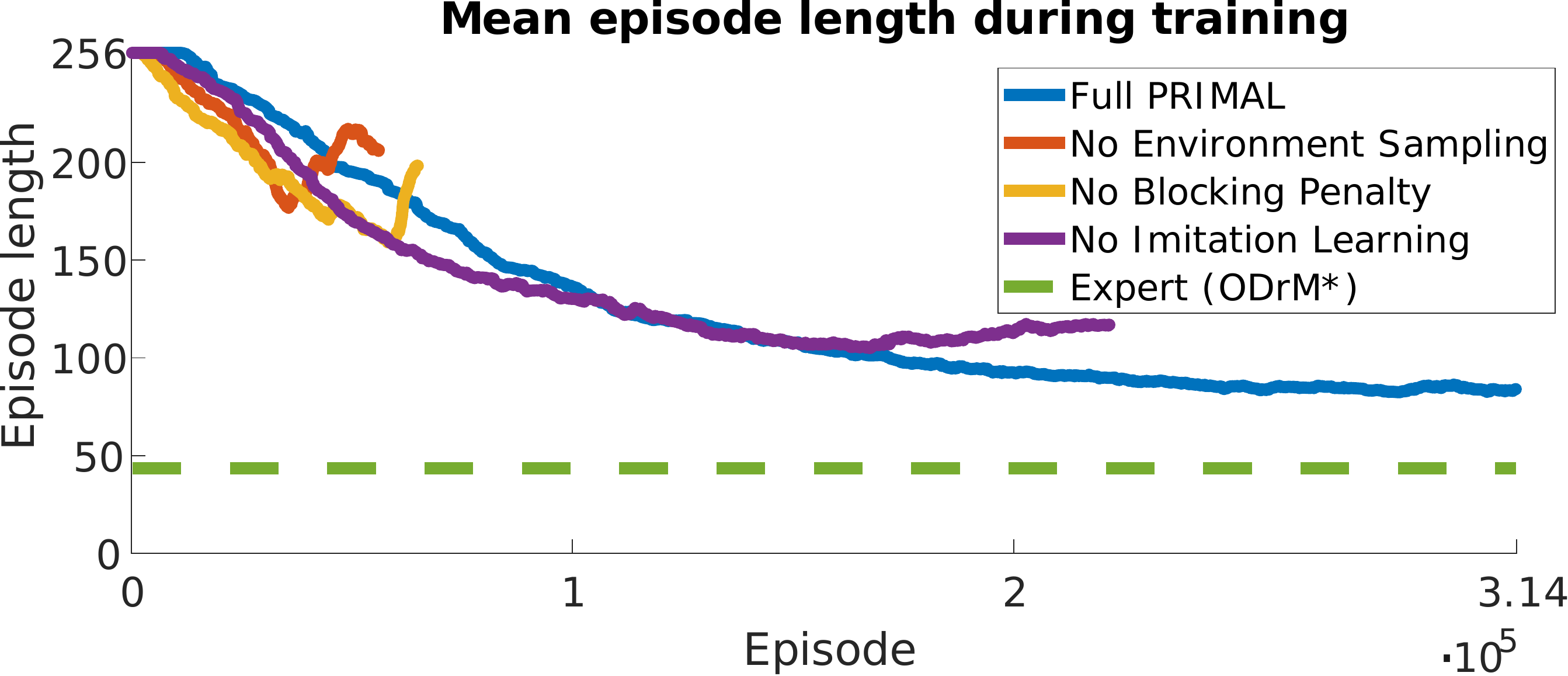}
\end{center}
\vspace{-0.6cm}
\caption{Mean episode length during training, lower is better.
The dotted line shows the baseline, obtained from the expert ODrM* planner.
When we remove either environment sampling, the blocking penalties, or imitation learning from our approach, the policy converges to a worse solution.}
\label{RAL2018-fig:ablation}
\vspace{-0.4cm}
\end{figure}

To successfully teach agents collaborative behavior, we rely on three methods: applying a penalty for encouraging other agents' movement (called the ``blocking penalty''), using expert demonstrations during training, and tailoring the random environments during training to expose agents to more difficult cluttered scenarios.
We emphasize that, without all three methods, the learning process is either unstable (no learning) or converges to a worse policy than with all three, as is apparent in Fig.~\ref{RAL2018-fig:ablation}.

\subsubsection{Blocking Penalty}
\label{RAL2018-learning-blocking}

First, we augment the reward function shown in Table~\ref{RAL2018-tab:rewardFunction} with a sharp penalty ($-2$ in practice) if an agent decides to stay on goal while preventing another agent from reaching its goal.
The intuition behind this reward is to provide an incentive for agents to leave their goals, offsetting the (selfish) local maximum agents experience while resting on goal. 
Our definition of blocking includes cases where an agent is not just preventing another agent from reaching its goal, but also cases where an agent delays another agent significantly (in practice, by $10$ or more steps to match the size of the agents' FOV).
This looser definition of blocking is necessary because of the agents' small FOV.
Although an alternate route might exists around the agent in larger worlds, it is illogical to move around the agent when coordination could lead to shorter a path, especially if the alternate route lies outside the agent's FOV (and therefore is uncertain). 

We use standard $A^*$ to determine the length of an agent's path from its current position to its goal and then that of its path when each one of the other agents is removed from the world.
If the second path is shorter than the first one by more than $10$ steps, that other agent is considered blocking.
The ``blocking'' output of the network is trained to predict when an agent is blocking others, to implicitly provide the agent with an ``explanation'' of the extra penalty it will incur in this case.

\subsubsection{Combining RL and IL}
\label{RAL2018-learning-ILRLcombination}

\begin{figure}[t]
\vspace{0.2cm}
\begin{center}
\includegraphics[width=\linewidth]{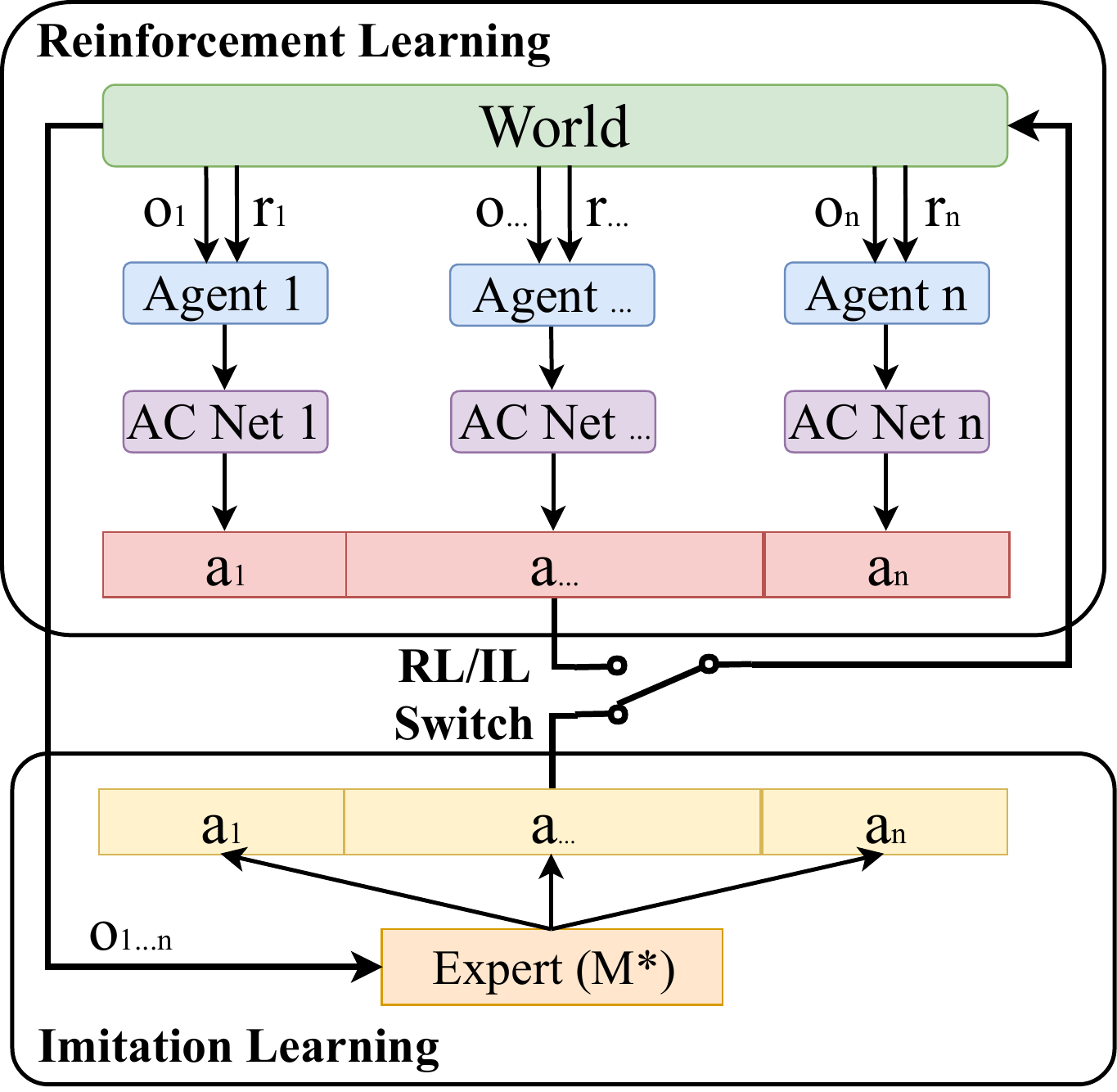}
\end{center}
\vspace{-0.55cm}
\caption{Structure of our hybrid RL/IL approach.
In the beginning of each episode, a random draw determines whether the episode will be RL- or IL-based, and the ``switch'' (in the middle) is set accordingly.
For the RL-based learning, at each timestep, each agent ($1,..,n$) draws its observation $o_i$ and reward $r_i$ for its previous action from the world (learning environment) and uses the observation to select an action $a_i$ via its own copy of the neural network.
The actions of different agents are executed sequentially in a random order.
Since agents often push and pull weights from a common, shared neural network, they ultimately share the same weights in their individual nets.
For the IL-based learning, an expert centralized planner coordinates all agents during the episode, whose behavior the agents learn to imitate, allowing them to learn coordinated behaviors.}
\label{RAL2018-fig:fullSystem}
\vspace{-0.3cm}
\end{figure}

Second, combining RL and IL has been shown to lead to faster, more stable training as well as higher-quality solutions in robot manipulation~\cite{nair2017overcoming,dextrousManipulation,gao2018reinforcement}.
These advantages are likely due to the fact that IL can help to quickly identify high-quality regions of the agent’s state-action space, while RL can further improve the policy by freely exploring these regions.
% Inspired by these works, we here let agents observe a demonstration of an optimal plan for some of the training episodes.
In our work, we randomly select in the beginning of each episode whether it will involve RL or IL (thus setting the central switch in the middle of Fig.~\ref{RAL2018-fig:fullSystem}).
Such demonstrations are generated dynamically by relying on the centralized planner ODrM*~\cite{wagner2015subdimensional} (with $\epsilon = 2$).
A trajectory of observations and actions $T\in (\mathcal{O}\times\mathcal{A})^n$ is obtained for each agent, and we minimize the behavior cloning loss:

\vspace{-0.5cm}
\begin{equation}
L_{bc}=-\frac{1}{T} \sum_{t=0}^T \log(P(a_t | \pi, o_t; \theta)).
\label{RAL2018-eq:lossFunction}
\end{equation}
\vspace{-0.37cm}

Our implementation deviates from~\cite{nair2017overcoming,vecerik2017leveraging} in that we combine off-policy behavior cloning with on-policy actor-critic learning, rather than with off-policy deep deterministic policy gradient.
We explored this approach since we can cheaply generate expert demonstrations online in the beginning of a new training episode, as opposed to other work where learning agents only have access to a finite set of pre-recorded expert trajectories.
The heuristic used in ODrM* inherently helps generate high-quality paths with respect to our reward structure (Table~\ref{RAL2018-tab:rewardFunction}), where agents move to their goals as quickly as possible (while avoiding collisions) and rest on it.
% The base of our reward structure is very similar to the heuristic used in ODrM*, with small additions centered around exploration and agent interactions.
Therefore, the RL/IL gradients are naturally coherent, thus avoiding oscillations in the learning process.

Leveraging demonstrations is a necessary component of our system: without it, learning progresses far slower and converges to a significantly worse solution.
However, we experimented with various IL proportions ($10$-$50\%$ by increments of $10\%$) and observed that the RL/IL ratio does not seem to affect the performance of the trained policy by much.
Finally, although we could use dynamic methods such as $\mbox{DA}_{\mbox{GGER}}$~\cite{ross2011reduction} or confident inference~\cite{chernova2007confidence} because of the availability of a real-time planner, we chose to use behavior cloning because of its simplicity and ease of implementation.
It is unclear whether using such methods would lead to a performance increase, and will be the subject of future works.

\subsubsection{Environment Sampling}
\label{RAL2018-learning-environmentSamp-ling}

Finally, during training, we randomize both the sizes and obstacle densities of worlds in the beginning of each episode.
We found that uniformly sampling the size and densities of worlds did not expose the agents to enough situations in which coordination is necessary because of the relative sparsity of agent-agent interactions.
We therefore sample both the size and the obstacle density from a distribution that favors smaller and denser environments, forcing the agents to learn coordination since they experience agent-agent interactions more often.

%%%%%%%%%%%%%%%%%%%%%%%%%%%%%%%%%%%%%%%%%%%%%%%%%%%%%%%%%%%%%%%%%%%%%%%%%%%%%%%%

\subsection{Training Details}
\label{RAL2018-learning-training}

\subsubsection{Environment}
\label{RAL2018-learning-environment}

The size of the square environment is randomly selected in the beginning of each episode to be either $10$, $40$, or $70$, with a probability distribution that makes $10$-sized worlds twice as likely.
The obstacle density is randomly selected from a triangular distribution between $0$ and $50\%$, with the peak centered at $33\%$.
The placement of obstacles, agents, and goals is uniformly at random across the environment, with the caveat that each agent had to be able to reach its goal.
That is, each agent is initially placed in the same connected region as its goal. %, disregarding other agents in the same region.
It is possible that agents train in impossible environments (e.g., two agents might be spawned in the same narrow connected region, each on the other's goal), although highly unlikely.
The actions of the agents are executed sequentially in a random order at each timestep to ensure that they have equal priority (i.e., race conditions are resolved randomly).

\subsubsection{Parameters}
\label{RAL2018-learning-parameters}

We use a discount factor ($\gamma$) of $0.95$, an episode length of $256$, and a batch size of $128$ so that up to two gradient updates are performed each episode per agent.
The probability of observing a demonstration is $50\%$ per episode.
We use the Nadam optimizer~\cite{dozat2016incorporating} with a learning rate beginning at $2 \cdot 10^{-5}$ and decaying proportionally to the inverse square root of episode count.
We train in $3$ independent environments with $8$ agents each, synchronizing agents in the same environment in the beginning of each step and allowing them to act in parallel.
Training was performed at the Pittsburgh Supercomputing Center (PSC)~\cite{Nystrom:2015:BUF:2792745.2792775} on $7$ cores of a Intel Xeon E5-2695 and one NVIDIA K80 GPU, and lasted around $20$ days.
The full code used to train agents is available at~\url{https://goo.gl/T627XD}.

%%%%%%%%%%%%%%%%%%%%%%%%%%%%%%%%%%%%%%%%%%%%%%%%%%%%%%%%%%%%%%%%%%%%%%%%%%%%%%%%
%%%%%%%%%%%%%%%%%%%%%%%%%%%%%%%%%%%%%%%%%%%%%%%%%%%%%%%%%%%%%%%%%%%%%%%%%%%%%%%%

\section{RESULTS}
\label{RAL2018-results}

In this section, we present the results of an extensive set of simulations comparing PRIMAL against state-of-the-art MAPF planners in gridworlds.
These tests are performed in environments with varying obstacle densities, grid sizes, and team sizes.
Finally, we present experimental results for a scenario featuring both physical and simulated robots planning paths online in an indoor factory mockup.

%%%%%%%%%%%%%%%%%%%%%%%%%%%%%%%%%%%%%%%%%%%%%%%%%%%%%%%%%%%%%%%%%%%%%%%%%%%%%%%%

\subsection{Comparison with Other MAPF Planners}
\label{RAL2018-results-gridworld}

For our experiments, we selected CBS~\cite{sharon2012conflict} as our optimal, centralized planner, ODrM*~\cite{wagner2015subdimensional} as suboptimal, centralized option (with inflation factors $\epsilon = 1.5$ and $\epsilon = 10$), and ORCA~\cite{van2011reciprocal} as fully-decoupled velocity planner.
Note that all other planners have access to the whole state of the system, whereas PRIMAL assumes that each agent only has partial observability of the system.
World sizes are $\{10,20,40,80,160\}$, densities $\{0,0.1,0.2,0.3\}$, and team sizes $\{4,8,...,1024\}$.
We placed no more than $32$ agents in $10$-sized worlds, no more than $128$ agents in $20$-sized worlds, and no more than $1024$ agents in $40$-sized worlds.

In our experiments, we compared the success rates of the different planners, that is whether they complete a given problem within a given amount of wall clock time or timesteps.
For CBS and ODrM*, we used a timeout of $300$s and $60$s, respectively, to match previous results~\cite{wagner2015subdimensional}.
We divided the timeout by $5$ for ODrM* because we used a $C++$ implementation which was experimentally measured to be about $5$ times faster than the previously used Python implementation.
For ORCA, we use a timeout of $60$s but terminate early when all agents are in a deadlock (defined as all agents being stuck for more than $120$s simulation time, which corresponds to $10$s physical time).
Finally, for PRIMAL, we let the agents plan individual paths for up to $256$ timesteps for $10$- to $40$-sized worlds, $384$ timesteps for $80$-sized worlds, and $512$ timesteps for $160$-sized worlds.
Experiments for the conventional planners were carried out on a single desktop computer, equipped with an AMD Threadripper $2990$WX with $64$ logical cores clocked at $4$Ghz and $64$Gb of RAM. % at $3200$MHz.
Experiments for PRIMAL were partially run on the same computer, which is also equipped with $3$ GPUs (NVIDIA Titan V, GTX $1080$Ti and $1070$Ti), as well as on a simple desktop with an Intel i$7$-$7700$K, $16$Gb RAM and an NVIDIA GTX $1070$.

Based on our results, we first notice that our approach performs extremely well in low obstacle densities, where agents can easily go around each other, but is easily outperformed in dense environments, where joint actions seem necessary for agents to reach their goals (which sometimes requires drastic path changes).
Similarly, but with significantly worse performance, ORCA cannot protect against deadlocks and performs very poorly in most scenarios involving more than $16$ agents and any obstacles, due to its fully-decoupled, reactive nature.
Second, we notice that, since our training involves worlds of varying sizes but a constant team size, agents are inherently exposed to a small variability in agent density within their FOV.
In our results, we observed that agents perform more poorly as the number of nearby agents increases in their FOV (small worlds, large teams), an effect we believe could be corrected by varying the team sizes during training.
This will be investigated in future works.
However, we expect traditional planners to generally outperform our approach in small ($10$-$20$-sized) worlds, even with larger teams.
Third, we notice that the paths generated by PRIMAL are sometimes more than twice as long as the paths of the other planners'.
However, other planners allow moves that the agents cannot take in our definition of the MAPF problem: agents can follow each other with no empty space between them, can swap around (similar to a runabout), etc.~\cite{wagner2015subdimensional}, which leads to shorter paths.
Additionally, visual inspection of the cases where PRIMAL generates longer paths shows that most agents move to their goals effectively, except for a few laggards.
Finally, since agents are never exposed to worlds larger than $70 \times 70$ during training, they seem to perform extremely poorly in larger worlds during testing ($\geq 80$-sized).
However, by capping the goal distance in the agents' state, PRIMAL's success rate in larger worlds can be drastically improved.
In the results presented here for $80$- and $160$-sized worlds, the distance to goal is capped at $75$ (empirically set) in the agents' state.
Example videos of near-optimal and severely sub-optimal plans for PRIMAL in various environments are available at~\url{https://goo.gl/T627XD}.

\begin{figure}[t]
\vspace{0.2cm}
\begin{center}
\includegraphics[width=0.96\linewidth]{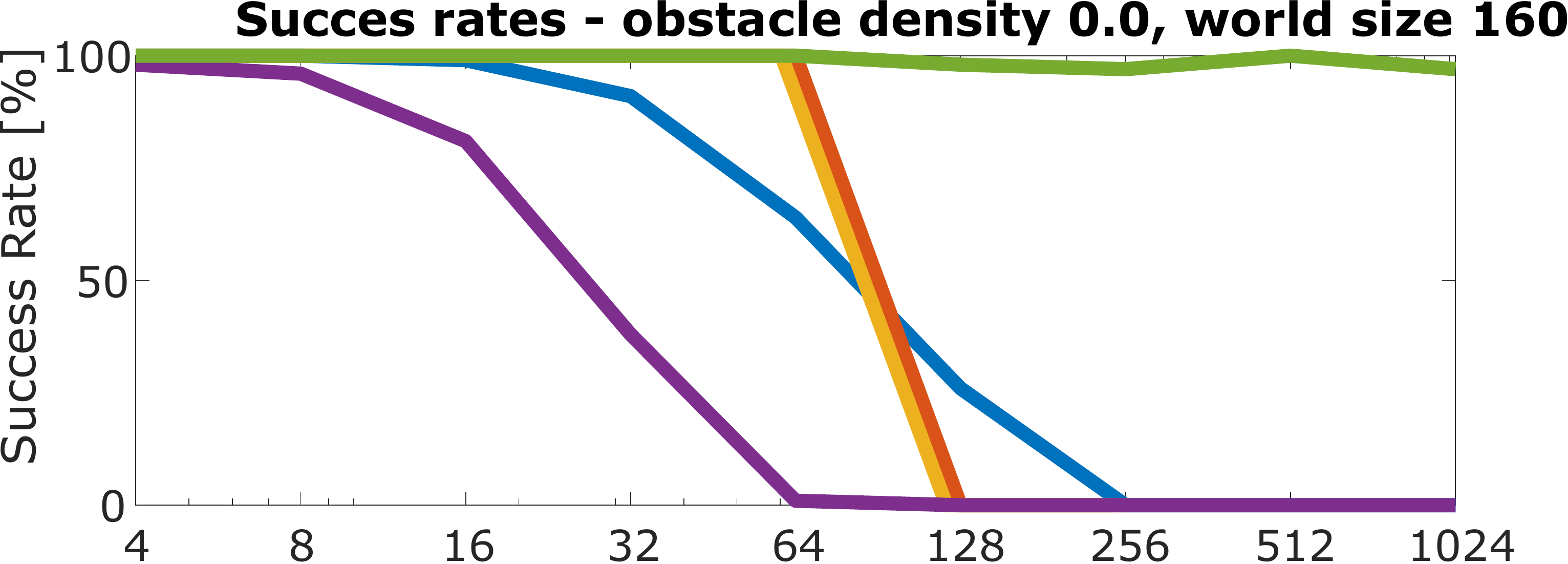} \\[0.15cm]
\includegraphics[width=0.96\linewidth]{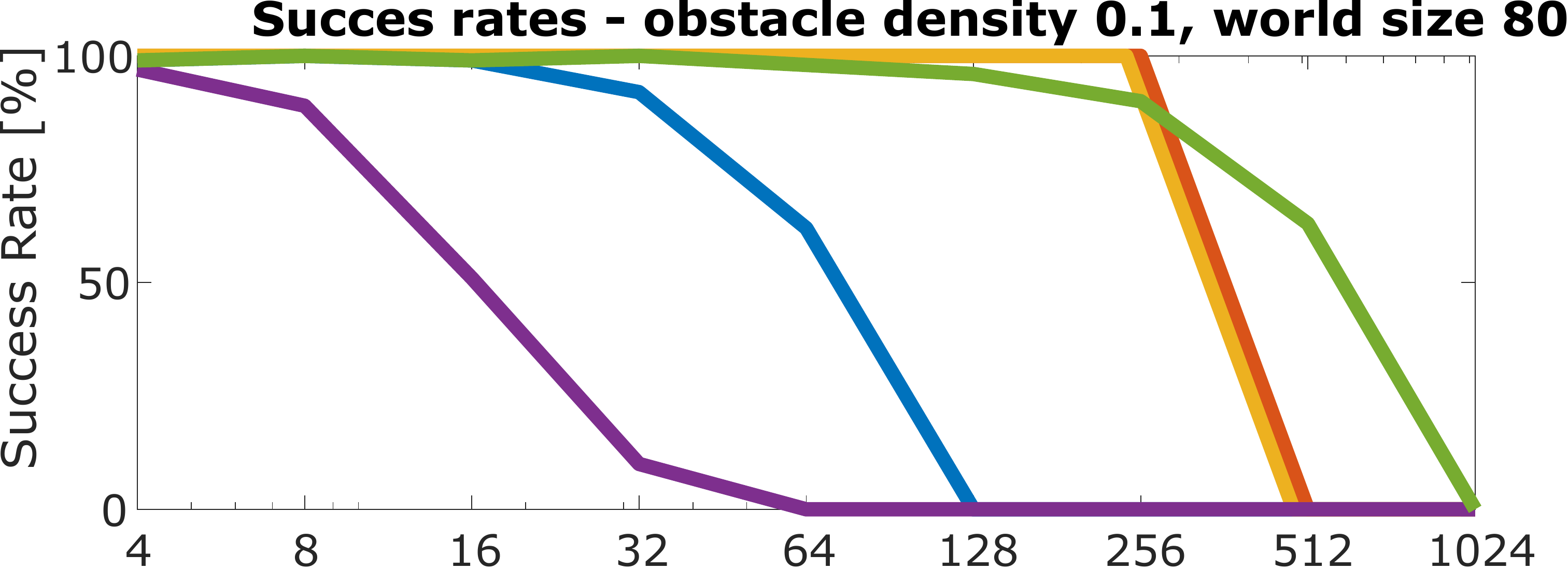}  \\[0.15cm]
\includegraphics[width=0.96\linewidth]{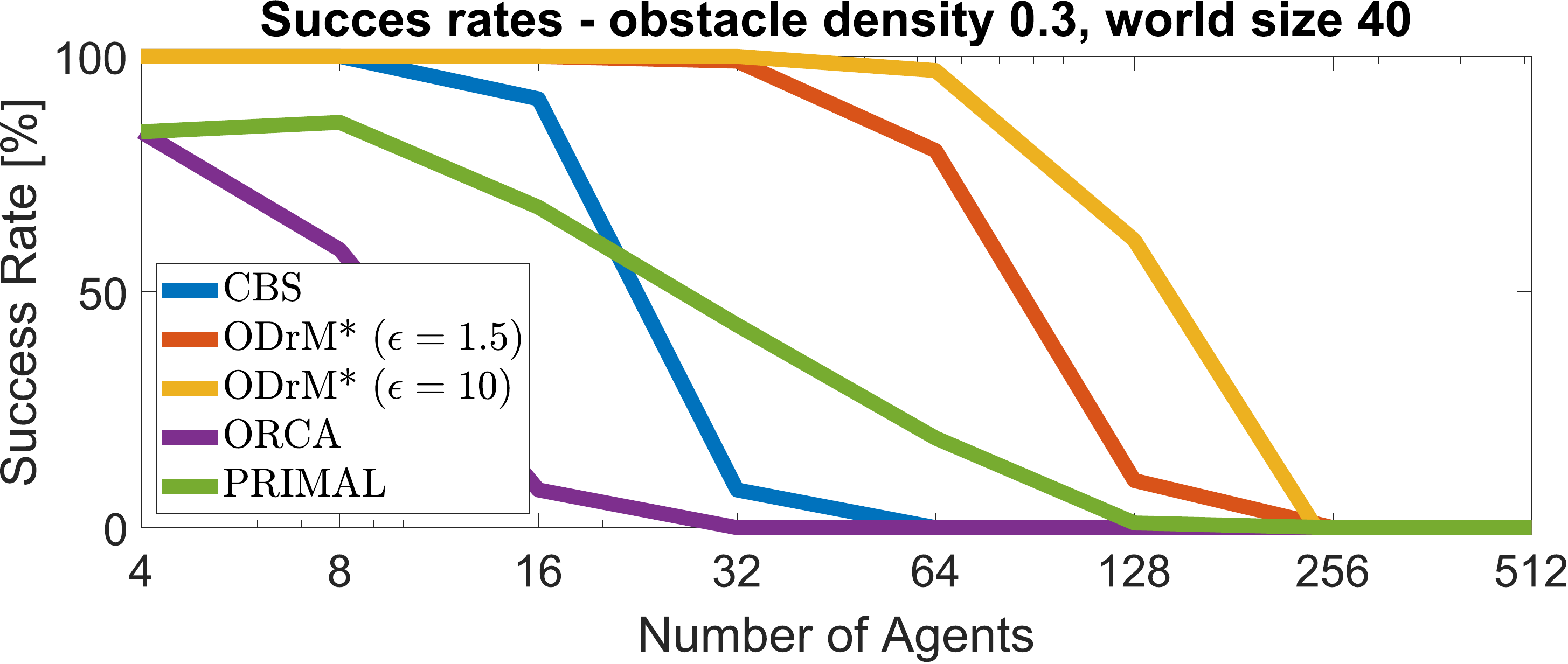}
\end{center}
\vspace{-0.6cm}
\caption{Success rates of the different planners in our three scenarios. PRIMAL outperforms all planners in the top obstacle-free world, slightly outperforms the others in low-obstacle-density worlds, and is strongly outperformed in the high-obstacle-density world.}
\label{RAL2018-fig:results}
% \vspace{-0.3cm}
\end{figure}

Due to space constraints, we choose to discuss the three main scenarios shown in Fig.~\ref{RAL2018-fig:results}: a case where PRIMAL strongly outperforms all other planners, one where PRIMAL slightly outperforms them, and one where PRIMAL struggles.
The complete set of results (for all team sizes, obstacles densities, and world sizes) can be found at~\url{https://goo.gl/APktNk} and contains the path lengths generated by the different planners as well as the planning times.
First, in a large world with no obstacles ($160 \times 160$), centralized planners especially struggle since the joint configuration space quickly grows to encompass all agents, making planning for more than $100$ agents very time-consuming.
PRIMAL, on the other hand, can easily deal with teams up to $1024$ agents, with a near-perfect success rate.
Second, in a medium-sized world with low obstacle density, the centralized planners can easily plan for a few hundred agents.
PRIMAL's success rate starts decreasing earlier than that of the other planners, but remains above $60\%$ for cases with $512$ agents, whereas all other planners perform poorly.
Third, in a smaller world that is very densely populated with obstacles, all planners can only handle up to $64$ agents, but PRIMAL starts to struggle past $8$ agents, whereas ODrM* can handle up to $64$ agents.
However, even when PRIMAL cannot finish a full problem, it usually manages to bring many agents to their goals quickly, with only a few failing to reach their goals.
At this point, a conventional planner could be used to complete the problem, which has become simple for a graph-based solver since most agents should remain motionless at their goals.
Future work will investigate the combination of PRIMAL with a complete planner to leverage the fast, decentralized planning of PRIMAL while guaranteeing completeness.

%%%%%%%%%%%%%%%%%%%%%%%%%%%%%%%%%%%%%%%%%%%%%%%%%%%%%%%%%%%%%%%%%%%%%%%%%%%%%%%%

\subsection{Experimental Validation}
\label{RAL2018-results-experiments}

\begin{figure}[t]
% \vspace{0.2cm}
\begin{center}
\includegraphics[width=0.9\linewidth]{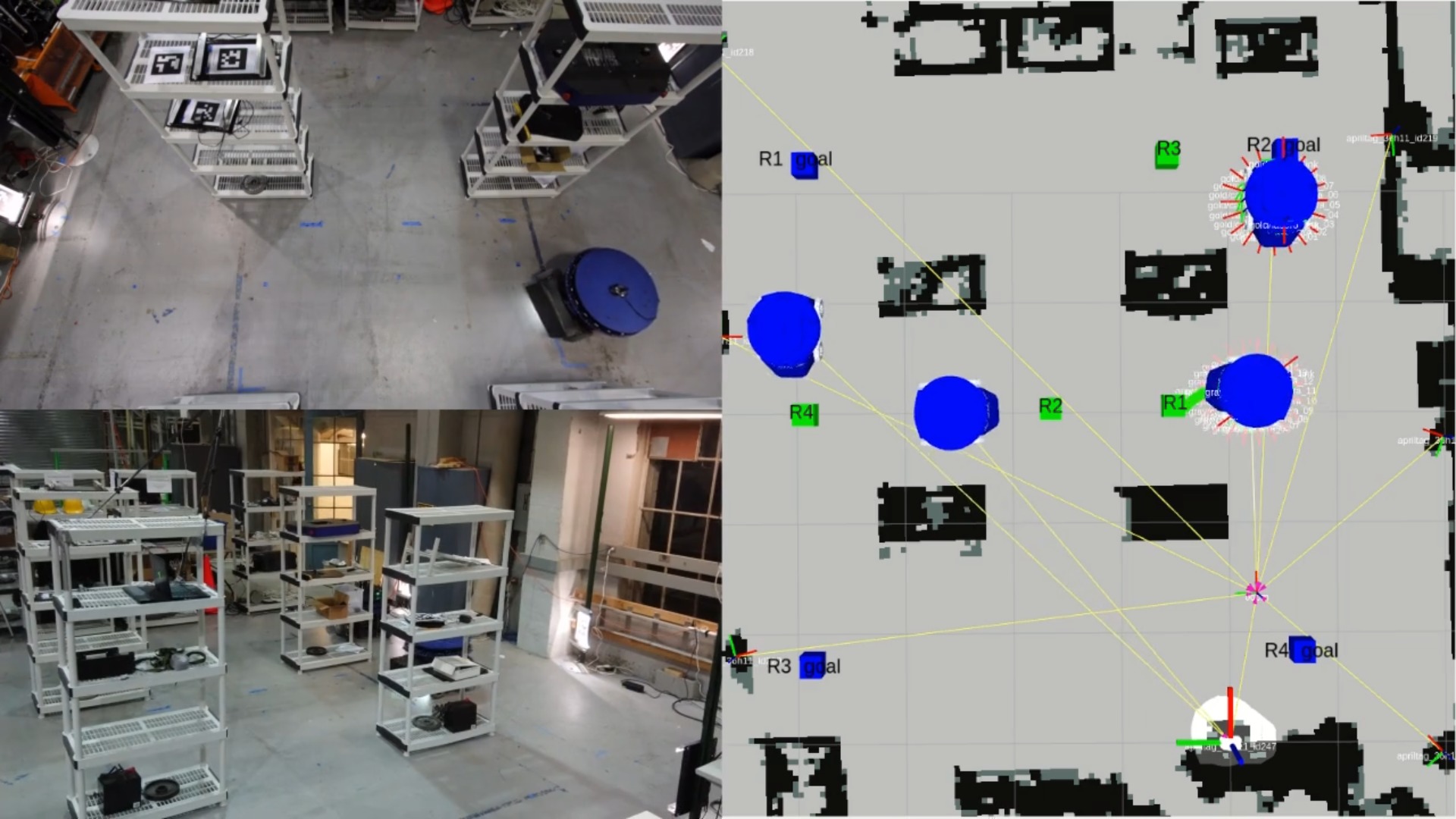}
\end{center}
\vspace{-0.52cm}
\caption{Snapshot of the physical and simulated robots evolving in the factory mockup. Left: overhead (top) and side (bottom) views of the mockup and robots. Right: visualization showing the obstacles (black solids), the robots (blue circles), their goals (blue squares), and current moves (green squares).}
\label{RAL2018-fig:hybridSim}
% \vspace{-0.3cm}
\end{figure}

We also implemented PRIMAL on a small fleet of autonomous ground vehicles (AGVs) evolving in a factory mockup.
In this hybrid system, two physical robots evolve alongside two (then, half-way through the experiment, three) simulated ones.
The physical robots have access to the position of simulated robots, and vice-versa, as they all plan their next actions online using our decentralized approach.
PRIMAL shows clear online capabilities, as the planning time per step and per agent is well below $0.1$s on a standard GPU (and well below $0.2$s on a CPU).
Fig.~\ref{RAL2018-fig:hybridSim} shows our factory mockup and simulation environment.
The full video is available at~\url{https://goo.gl/T627XD}.
% \vspace{-0.24cm}

% \begin{figure*}[t]
% %\vspace{0.2cm}
% \begin{center}
% \includegraphics[width=0.3\linewidth]{hybridSim01.jpg} \hspace{0.2cm}
% \includegraphics[width=0.3\linewidth]{hybridSim02.jpg} \hspace{0.2cm}
% \includegraphics[width=0.3\linewidth]{hybridSim03.jpg}
% \end{center}
% \vspace{-0.52cm}
% \caption{Successive frames of the physical and simulated robots exchanging positions in the factory mockup.}
% \label{RAL2018-fig:hybridSim}
% \vspace{-0.3cm}
% \end{figure*}

%%%%%%%%%%%%%%%%%%%%%%%%%%%%%%%%%%%%%%%%%%%%%%%%%%%%%%%%%%%%%%%%%%%%%%%%%%%%%%%%
%%%%%%%%%%%%%%%%%%%%%%%%%%%%%%%%%%%%%%%%%%%%%%%%%%%%%%%%%%%%%%%%%%%%%%%%%%%%%%%%

\section{CONCLUSION}
\label{RAL2018-conclusion}

In this paper, we presented PRIMAL, a new approach to multi-agent path finding, which relies on combining distributed reinforcement learning and imitation learning from a centralized expert planner.
Through an extensive set of experiments, we showed how PRIMAL scales to various team sizes, world sizes and obstacle densities, despite only giving agents access to local information about the world.
In low obstacle-density environments, we further showed how PRIMAL exhibits on-par performance, and even outperforms state-of-the-art MAPF planners in some cases, even though these have access to the whole state of the system.
Finally, we presented an example where we deployed PRIMAL on physical and simulated robots in a factory mockup, showing how robots can benefit from our online, local-information-based, decentralized MAPF approach.

Future work will focus on adapting our training procedure to factory-like environments, with low to medium obstacle density but where parts of the environment are very sparse and other parts highly-structured (such as corridors, aisles, etc.).
We also believe that extending our approach to receding-horizon planning, where agents plan ahead for several actions, may help to improve the performance of PRIMAL by teaching agents to explicitly coordinate their~paths.


%%%%%%%%%%%%%%%%%%%%%%%%%%%%%%%%%%%%%%%%%%%%%%%%%%%%%%%%%%%%%%%%%%%%%%%%%%%%%%%%
%%%%%%%%%%%%%%%%%%%%%%%%%%%%%%%%%%%%%%%%%%%%%%%%%%%%%%%%%%%%%%%%%%%%%%%%%%%%%%%%

\vspace{0.2cm}
\section*{ACKNOWLEDGMENTS}
\label{RAL2018-acknowledgments}
\noindent 
Detailed comments from anonymous referees contributed to the presentation and quality of this paper.
This research was supported by the CMU Manufacturing Futures Initiative, made possible by the Richard King Mellon Foundation.
Justin Kerr was a CMU SURF student, funded by the National Science Foundation (NSF).
The research at USC was supported by NSF grants 1409987, 1724392, 1817189, and 1837779.
This work used the Bridges system, supported by NSF grant ACI-1445606 at the Pittsburgh Supercomputing Center~\cite{Nystrom:2015:BUF:2792745.2792775}.

%%%%%%%%%%%%%%%%%%%%%%%%%%%%%%%%%%%%%%%%%%%%%%%%%%%%%%%%%%%%%%%%%%%%%%%%%%%%%%%%

\bibliographystyle{IEEEtran}
\bibliography{distributedRL_MAPF}

\end{document}